\documentclass[runningheads]{llncs}
\usepackage{latexsym}
\usepackage{booktabs}
\usepackage{graphicx}
\usepackage{amsmath}
\begin{document}
\title{Exploiting News Article Structure for Automatic Corpus Generation of Entailment Datasets}
\titlerunning{Exploiting News Article Structure for Automatic Corpus Generation}
%

\author{Jan Christian Blaise Cruz\inst{1} \and 
Jose Kristian Resabal\inst{2} \and
James Lin\inst{1} \and
Dan John Velasco\inst{1} \and 
Charibeth Cheng\inst{1}}
\authorrunning{J. Cruz et al.}
\institute{De La Salle University Manila, Taft Ave., Malate, 1004 Manila, Philippines \\
\email{\{jan\_christian\_cruz,james\_lin,dan\_velasco,charibeth.cheng\}@dlsu.edu.ph} \and 
University of the Philippines Diliman, Quezon Hall, 1101 Quezon City, Philippines \\
\email {jkresabal@up.edu.ph}
}


\maketitle              

\begin{abstract}
Transformers represent the state-of-the-art in Natural Language Processing (NLP) in recent years, proving effective even in tasks done in low-resource languages. While pretrained transformers for these languages can be made, it is challenging to measure their true performance and capacity due to the lack of hard benchmark datasets, as well as the difficulty and cost of producing them. In this paper, we present three contributions: First, we propose a methodology for automatically producing Natural Language Inference (NLI) benchmark datasets for low-resource languages using published news articles. Through this, we create and release NewsPH-NLI, the first sentence entailment benchmark dataset in the low-resource Filipino language. Second, we produce new pretrained transformers based on the ELECTRA technique to further alleviate the resource scarcity in Filipino, benchmarking them on our dataset against other commonly-used transfer learning techniques. Lastly, we perform analyses on transfer learning techniques to shed light on their true performance when operating in low-data domains through the use of degradation tests.

\end{abstract}

\section{Introduction}
In recent years, Transformers \cite{vaswani2017attention} have begun to represent the state-of-the-art not only in common NLP tasks where they have cemented their reputation, but also in the context of tasks within low-resource languages. Using Transformers, advancements have been done in various low-resource tasks, including low-resource translation \cite{currey2019incorporating,murray2019auto}, classification \cite{myagmar2019cross,cruz2020localization}, summarization \cite{khandelwal2019sample}, and many more.

Transformers and transfer learning techniques in general owe their wide adaptation in low-resource language tasks to the existence of abundant unlabeled corpora available. While labeled datasets may be scarce to perform tasks in these languages, unlabeled text is usually freely available and can be scraped from various sources such as Wikipedia, news sites, book repositories, and many more. Pretraining allows transfer learning techniques to leverage learned priors from this unlabeled text to robustly perform downstream tasks even when there is little task-specific data to learn from \cite{howard2018universal,cruz2019evaluating}.

However, while it is possible to produce large pretrained models for transfer learning in low-resource languages, there is a challenge in properly gauging their performance in low-resource tasks. Most, if not all Transformers that are pretrained and released open-source are evaluated with large, commonly-used datasets. In low-resource languages, these datasets may not exist. Due to this, it is often hard to properly benchmark a model's true performance when operating in low-data domains.

While it is possible to remedy this by constructing hard datasets in these languages, added concerns have to be addressed. 

Dataset construction is slow and cost-prohibitive. For hard tasks such as various natural language inference and understanding tasks, datasets are usually sized around 500,000 samples and more \cite{bowman2015large,conneau2018xnli,williams2017broad}. This would entail a large enough budget to hire annotators to write text samples, and a different set of annotators to write labels. This process is also slow and may take months to finish. In that span of time, stronger techniques may have been created that require more difficult datasets to accurately assess them. In addition, once the dataset has been solved, harder datasets are needed to properly gauge further, succeeding methods.

This creates a need for a method to produce benchmark datasets for low-resource languages that is quick and cost effective, while still capable of generating tasks that are challenging for high-capacity models such as Transformers.

In this paper, we present the following contributions:
\begin{itemize}
\item We propose an automatic method to generate Natural Language Inference (NLI) benchmark datasets from a corpus of news articles.
\item We release \textbf{NewsPH-NLI}, the first sentence entailment benchmark dataset in the low-resource Filipino language, created using the method we propose.
\item We produce pretrained Transformers based on the ELECTRA pretraining scheme to further alleviate resource scarcity in Filipino.
\item We perform benchmarks and analyses on commonly-used transfer learning techniques to properly and accurately gauge their true performance in low-data domains.
\end{itemize}

Our method has a number of advantages. First, since our method is automatic, it evades the issue of time and cost. This also allows datasets created this way to be updated regularly as news is released everyday. Second, given that news is freely available and published online even in low-resource languages, text data for producing benchmark datasets will be easy to source. Lastly, given that we generate sentence entailment tasks within the domain of news, our method will produce sufficiently challenging datasets to properly gauge the performance of large Transformers.

\section{Methodology}
In this section, we outline our experimental setups and methodology. First, we describe our proposed methodology for producing benchmark NLI datasets in any language, using our NewsPH-NLI dataset as an example. Second, we outline the creation of ELECTRA models in Filipino. Lastly, we outline our methodology for analysis using degradation tests.

\subsection{NLI Datasets from News Articles}
The creation of large datasets for NLI is often difficult, time-consuming, and cost-prohibitive. It may also be not be feasible in low-data and low-resource domains due to the lack of pre-encoded text data that can immediately be annotated.

\begin{figure}[htp]
    \centering
    \includegraphics[width=0.48\textwidth]{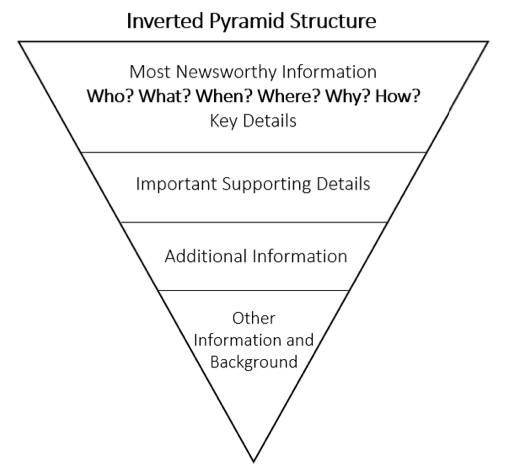}
    \caption{\label{fig:inverted-pyramid} 
    Inverted pyramid structure of straight news articles. The most important facts are at the top with succeeding paragraphs containing facts of less and less importance. The structure of news articles makes succeeding paragraphs build up on the information of prior paragraphs. Figure taken from \cite{norambuena2020evaluating}}
\end{figure}

We propose the use of news articles for automatically creating benchmark datasets for NLI because of two reasons. First, news articles commonly use single-sentence paragraphing, meaning every paragraph in a news article is limited to a single sentence \cite{hoey2008beginning,hinds1977paragraph}. Second, straight news articles follow the ``inverted pyramid'' structure, where every succeeding paragraph builds upon the premise of those that came before it, with the most important information on top and the least important towards the end \cite{lamble2013news,canavilhas2007web,po2003news}. A figure illustrating the inverted pyramid can be found in Figure \ref{fig:inverted-pyramid}.


Due to the inverted pyramid structure, we can assume that every succeeding paragraph (our ``hypothesis'') entails the paragraph preceeding it (our ``premise''). This can be exploited to produce multiple samples of entailments from a single news article. This way, using a corpus of straight news articles, we can produce a large number of samples of entailments to make up an NLI corpus.

Contradictions, on the other hand, are more difficult to produce. To automatically make contradiction samples, we first randomly sample two news articles from the pool of collected news articles, then randomly sample one paragraph from each article to serve as our premise-hypothesis pair. To ensure that the produced pair is a contradiction, we must first make sure that the two randomly-sampled articles have two different topics. To do this, we train a Doc2Vec \cite{le2014distributed} model on all of the collected news articles. Afterwards, we then cluster the most similar articles. When sampling articles for contradictions, we sample from two different clusters to ensure that the topics are different.

One limitation of our proposed methodology is that it can only generate entailments and contradictions, as ``neutral'' premise-hypothesis pairs can only be obtained through manual annotation by humans. This lack of a third label makes the generated datasets easier as compared to standard NLI datasets with three labels. While a 2-label classification task is easier than a 3-label classification task, the generated dataset will still be harder than a standard single-sentence classification problem (like sentiment classification) as the model will have to be able to encode inter-dependent information between the sentence pairs.

In addition, there is a chance that an auto-generated dataset will have errors that can only be identified when checked and studied by human annotators. As the goal of the research is to produce a dataset with little-to-no human supervision nor annotation, this human-based check is not done. Correctness is instead ensured by thorough testing of the topic clustering model.

\subsection{NewsPH-NLI}
Using our proposed methodology, we automatically generate an NLI benchmark dataset in Filipino we call the NewsPH-NLI dataset. 

To create the dataset, we scrape news articles from all major Philippine news sites online. We collect a total of 229,571 straight news articles, which we then lightly preprocess to remove extraneous unicode characters and correct minimal misspellings. No further preprocessing is done to preserve information in the data.

We then use our proposed methodology. First, we create a Doc2Vec model (via the Gensim\footnote{https://radimrehurek.com/gensim/} package) on our collected news corpus, using Annoy\footnote{https://github.com/spotify/annoy} as an indexer. We remove Tagalog stopwords and use TF-IDF to filter the functions words (e.g. ``ng`` and ``nang``) as these create noise. In testing, without the use of stopword removal and TF-IDF filtering, clustering was difficult as most articles were embedded closely due to their common usage of stopwords and function words. After producing the Doc2Vec embeddings, we then cluster, comparing two articles via the cosine similarity of the mean of their projected vectors. We consider two articles to be dissimilar if their cosine similarity is less than 0.65.

After clustering, we then take entailments by running through each article iteratively, and produce contradictions by sampling from two randomly chosen clusters. We shuffle the final set and randomly sample 600,000 premise-hypothesis pairs to be part of the final dataset. We set this size for our dataset in order to follow the size of the widely-used SNLI dataset \cite{bowman2015large}.

From the full generated dataset, 420,000 of which form the training set, while the remaining 80,000 are split evenly to produce the validation and test sets. To generate the splits, we first sample 300,000 of both entailments and contradictions using our methodology, shuffle the set, then split them accordingly into training, validation, and testing sets.

\subsection{ELECTRA Pretraining}
We alleviate the resource scarcity of the Filipino language by producing pretrained transformers. We chose the ELECTRA \cite{clark2020electra} pretraining method because of the data efficiency of its pretraining task. While a large corpus of unlabeled text is available in Filipino, this consolidated corpus is still far smaller than the ones commonly used to pretrain English models. ELECTRA poses an advantage over the widely-used BERT \cite{devlin2018bert} in its ability to use pretraining data more efficiently, as BERT only uses 15\% of the training data for masked language modeling per epoch, leading to data inefficiency. We surmise that this increased data efficiency will provide improvements for tasks in low-resource languages.

We produce four ELECTRA models: a cased and uncased model in the base size (12 layers, 768 hidden units, 12 attention heads), and a cased and uncased model in the small size (12 layers, 256 hidden units, 4 attention heads). All our models accept a maximum sequence length of 512.

Our models are pretrained using the WikiText-TL-39 dataset \cite{cruz2019evaluating}, producing a SentencePiece\footnote{https://github.com/google/sentencepiece} vocabulary of 320,000 subwords. We train the small models with a learning rate of 5e-4, batch size of 128, and a generator hidden size 25\% of the discriminator hidden size. For the base models, we train with a learning rate of 2e-4, batch size of 256, and a generator hidden size 33\% of the discriminator hidden size. Models are pretrained using the Adam \cite{kingma2014adam} optimizer. We pretrain for a total of 1 million steps for the base models and 766,000 steps for the small models, using the first 10\% of the total steps for linear learning rate warmup.

Pretraining was done using Tensor Processing Unit (TPU) v3 machines on Google Cloud Platform, with small models finishing in four days and base variants finishing in nine days.


\subsection{Benchmarking}
We then finetune to set initial benchmarks on the NewsPH-NLI using our ELECTRA models, comparing their performance against another Transformer-based finetuning technique (BERT) and an RNN-based finetuning technique (ULMFiT). For Filipino versions of the aforementioned benchmark models, we use Tagalog-BERT \cite{cruz2020establishing,cruz2020localization} and Tagalog-ULMFiT \cite{velasco2020pagsusuri}.

For finetuning, small variants of ELECTRA use a learning rate of 2e-4. Base variants of both ELECTRA and BERT use a learning rate of 5e-5. All transformers were finetuned on the dataset for a total of 3 epochs using the Adam optimizer, using the first 10\% of the total steps for linear learning rate warmup. For transformers, the standard separator token \texttt{[SEP]} was used to convert sentence pairs into one single sequence.

ULMFiT follows a different finetuning protocol compared to the transformer models. We first preprocess the data using the FastAI \cite{howard2020fastai} tokenization scheme. Sentence-pairs are turned into one sequence by using a special \texttt{xxsep} token introduced in finetuning.

Finetuning was done in two stages: language model finetuning, and classifier finetuning. For language model finetuning, we first finetune the last layer for 1 epoch, leaving all other layers frozen, before unfreezing all layers and finetuning for two epochs. We use a learning rate of 5e-2. For classifier finetuning, we perform 5 epochs of finetuning, performing gradual unfreezing \cite{howard2018universal} while reducing the learning rate from 1e-2 per epoch by a factor of 2. All experiments with ULMFiT also used discriminative learning rates \cite{howard2018universal} and cyclic learning rate schedules \cite{smith2017cyclical}.

Finetuning, testing, and all other experiments were done on machines with NVIDIA Tesla P100 GPUs. For small ELECTRA models, finetuning on the full dataset takes three hours to finish. For base ELECTRA and BERT variants, full finetuning finishes in five hours. For ULMFiT, it takes two hours

\subsection{Degradation Tests}
To further investigate the capacity and performance of these models especially when operating in low-data environments, we run a number of degradation tests \cite{cruz2020establishing}.

Simply put, we reduce the amount of training data to a certain \textbf{data percentage} (p\%) of the full dataset while keeping the validation and testing data sizes constant, then proceed to finetune a model. For each model, we perform degradation tests at four different data percentages: 50\%, 30\%, 10\%, and 1\%.

For each degradation test, we log the test loss and test accuracy. In addition, we take the \textbf{accuracy degradation}, which is described as:
$$
AD_{p\%} = Acc_{100\%} - Acc_{p\%}
$$
where $Acc_{100\%}$ refers to the accuracy of the model when finetuned on the full dataset, and $Acc_{p\%}$ refers to the accuracy of the model when finetuned on $p\%$ of the dataset. We also take the \textbf{degradation percentage}, which is described as:
$$
DP_{p\%} = AD_{p\%} / Acc_{100\%} \times 100
$$
where $AD_{p\%}$ is the accuracy degradation of the model when finetuned at $p\%$ data percentage. The degradation percentage measures how much of the full performance of a model is lost when trained with less data, at a certain data percentage $p\%$.

To compare which models are more robust to performance degradation in low-data domains, we also measure \textbf{degradation speed}, which we define as the standard deviation between the degradation percentages of a model for the 50\%, 30\%, 10\%, and 1\% setup. When the degradation percentages are more spread out, this indicates that the model degrades faster as the number of training data is reduced.

We perform degradation tests as a form of ``stress test'' to gauge the performance and effectiveness of models when forced to work in low-data domains. Most models in published literature show results as tested in environments with abundant data. While this is an effective way to compare performance against other models tested in a similar manner, it is not representative of a model's actual performance when adapting to low-data domains, especially with low-resource languages.

\section{Results and Discussion}

\subsection{Finetuning Results}

Finetuning results show that ELECTRA outperforms both the Transformer baseline (BERT) and the RNN baseline (ULMFiT). The best ELECTRA model (Small Uncased) outperforms the best BERT model (Base Cased) by +3.75\% accuracy, and outperforms the ULMFiT model by +3.63\%.

The ELECTRA models outperformed the BERT models on average by 3.01\% accuracy (average ELECTRA performance being 92.17\% while average BERT performance is only 89.16\%). We hypothesize that the ELECTRA models perform better than the BERT, with the small variants performing better than their larger BERT counterparts despite the size and capacity difference, due to the pretraining scheme. ELECTRA leverages pretraining data in a more data efficient way, using all of the training data per batch to train the model. This is opposed to BERT's (particularly masked language modeling's) inefficient use of pretraining data, using only 15\% of each batch to train the model. Since our pretraining dataset is considerably smaller than most common English pretraining datasets (39 million words in WikiText-TL-39 vs 2,500 million words in the Bookcorpus dataset), a pretraining scheme that uses data more efficiently will be able to learn more effectively.

Difference in performance among the ELECTRA variants is marginal at best, with the difference in accuracy between the best ELECTRA model (Small Uncased) and the weakest one (Base Cased) being only 1.22\%. An interesting observation is that the small variants both outperform their base variants, albeit marginally. The small uncased model outperforms the base uncased model by 1.34\%, while the small cased model outperforms the base cased model by 0.51\%. We hypothesize that this is due to the small models being easier to train, given that there are less parameters to consider.

While the small variants outperform their base variants on the full dataset, we hypothesize that the base models have an advantage in settings where there is less data to learn from, since they have more effective capacity. We verify this through our use of degradation tests shown in the next subsection.


\begin{table*}
\centering\begin{tabular}{lccccc}
\toprule
                 Model & Val. Loss & Val. Acc. & Test Loss & Test Acc. &  Seed \\
\midrule
    ELECTRA Tagalog Base Cased &    0.2646 &    91.74\% &    0.2619 &    91.76\% &  4567 \\
  ELECTRA Tagalog Base Uncased &    0.2502 &    91.98\% &    0.2581 &    91.66\% &  4567 \\
   ELECTRA Tagalog Small Cased &    0.1931 &    92.58\% &    0.1959 &    92.27\% &  1439 \\
 ELECTRA Tagalog Small Uncased &    0.1859 &    92.96\% &    0.1894 &    93.00\% &    45 \\
 \midrule
       BERT Tagalog Base Cased &    0.3225 &    88.81\% &    0.3088 &    89.25\% &  1111 \\
     BERT Tagalog Base Uncased &    0.3236 &    89.04\% &    0.3257 &    89.06\% &  6235 \\
                ULMFiT Tagalog &    0.2685 &    89.11\% &    0.2589 &    89.37\% &    42 \\
\bottomrule
\end{tabular}
\caption{\label{tab:main-results}
Final Finetuning Results. The best ELECTRA model (Small Uncased) outperforms the best BERT model (Base Cased) by +3.75\% and the ULMFiT model by +3.63\%. An interesting observation is that the small ELECTRA models perform marginally better than their base counterparts. We also report the random seed used in our experiments for reproducibility with our released code.}
\end{table*}

A table summarizing the finetuning results can be found in Table \ref{tab:main-results}.

\subsection{Degradation Tests}

In total, we perform four degradation tests per model variant, for a total of 28 degradation tests. Each model is finetuned with a fraction of the entire NewsPH-NLI dataset (50\%, 30\%, 10\%, and 1\%), with the resulting performance compared against the performance of the same model when finetuned with the full dataset. A summary of all degradation tests can be found in Table \ref{tab:degradation-tests}.


\begin{table*}
\centering\begin{tabular}{llccccc}
\toprule
                 Model & Data \% & Test Loss & Test Acc & Acc. Deg. &  Deg. \% & Degradation Speed \\
\midrule
    ELECTRA Tagalog  &   100\% &    0.2619 &   91.76\% &          &       \\
    Base Cased         &    50\% &    0.3184 &   90.56\% &      -1.20 &   1.31\% \\
                       &    30\% &    0.3769 &   88.85\% &      -2.91 &   3.17\% & 4.47 \\
                       &    10\% &    0.4467 &   86.23\% &      -5.53 &   6.03\% \\
                       &     1\% &    0.5046 &   79.78\% &     -11.98 &  13.06\% \\
\midrule
  ELECTRA Tagalog  &   100\% &    0.2581 &   91.66\% &           &        \\
  Base Uncased         &    50\% &    0.2920 &   90.85\% &      -0.81 &   0.88\% \\
                       &    30\% &    0.3333 &   89.21\% &      -2.45 &   2.67\% & 4.77 \\
                       &    10\% &    0.4041 &   87.20\% &      -4.46 &   4.87\% \\
                       &     1\% &    0.5300 &   79.43\% &     -12.23 &  13.34\% \\
\midrule
   ELECTRA Tagalog  &   100\% &    0.1959 &   92.27\% &           &        \\
   Small Cased        &    50\% &    0.2260 &   91.56\% &      -0.71 &   0.77\% \\
                       &    30\% &    0.2504 &   90.13\% &      -2.14 &   2.32\% & 5.69 \\
                       &    10\% &    0.3075 &   87.66\% &      -4.61 &   5.00\% \\
                       &     1\% &    0.4873 &   78.09\% &     -14.18 &  15.37\% \\
\midrule
 ELECTRA Tagalog  &   100\% &    0.1894 &   93.00\% &           &        \\
 Small Uncased         &    50\% &    0.2154 &   91.97\% &      -1.03 &   1.11\% \\
                       &    30\% &    0.2439 &   90.86\% &      -2.14 &   2.30\% & 6.92 \\
                       &    10\% &    0.2963 &   88.77\% &      -4.23 &   4.55\% \\
                       &     1\% &    0.5303 &   75.91\% &     -17.09 &  18.38\% \\
\midrule
\midrule
       BERT Tagalog  &   100\% &    0.3088 &   89.25\% &           &        \\
       Base Cased     &    50\% &    0.3800 &   87.09\% &      -2.16 &   2.42\% \\
                       &    30\% &    0.4394 &   86.25\% &      -3.00 &   3.36\% & 2.49 \\
                       &    10\% &    0.5046 &   84.15\% &      -5.10 &   5.71\% \\
                       &     1\% &    0.5285 &   81.33\% &      -7.92 &   8.87\% \\
\midrule
     BERT Tagalog  &   100\% &    0.3257 &   89.06\% &           &        \\
     Base Uncased       &    50\% &    0.4126 &   87.15\% &      -1.91 &   2.14\% \\
                       &    30\% &    0.4434 &   86.04\% &      -3.02 &   3.39\% & 2.57 \\
                       &    10\% &    0.5232 &   83.65\% &      -5.41 &   6.07\% \\
                       &     1\% &    0.5672 &   81.21\% &      -7.85 &   8.81\% \\
\midrule
                ULMFiT Tagalog &   100\% &    0.2589 &   89.37\% &           &        \\
                       &    50\% &    0.3093 &   86.82\% &      -2.55 &   2.85\% \\
                       &    30\% &    0.3699 &   84.39\% &      -4.98 &   5.57\% & 8.33 \\
                       &    10\% &    0.4840 &   79.07\% &     -10.30 &  11.53\% \\
                       &     1\% &    0.8140 &   67.50\% &     -21.87 &  24.47\% \\
\bottomrule
\end{tabular}
\caption{\label{tab:degradation-tests}
Degradation Test Results. ``Acc. Deg.'' refers to Accuracy Degradation, the difference between the performance of the model when trained with the full dataset and when trained with a smaller Data \%. ``Deg. \%'' refers to Degradation Percentage, the percentage of the performance of the model when trained with the full dataset that is lost when finetuned with a smaller Data \%. Degradation speed is the standard deviation of a model's Degradation Percentages, lower is better.}
\end{table*}

As we start to reduce the training data to 50\%, the ELECTRA models remain more resilient to performance degradation compared to the BERT models and ULMFiT. We hypothesize this to be due to the more effective means of imparting learned priors to the Transformer by its data-efficient pretraining scheme. At the 50\% data percentage, ELECTRA has only degraded by 1.02\% on average, while BERT and ULMFiT has degraded by 2.28\% and 2.85\% on average, respectively. This trend is still evident at the 30\% data percentage mark, with ELECTRA degrading by 2.62\% on average, while BERT and ULMFiT degrade by 3.38\% and 5.57\% on average, respectively.

The trend begins to shift as we approach settings with even less data. As the training data is reduced to 10\% of the original (42,000 examples), we see that ELECTRA starts to begin degrading faster, while BERT degrades at about the same rate. ELECTRA has degraded by 5.11\% on average. Meanwhile, BERT degrades by 5.89\% on average, which is only minimally larger than ELECTRA's degradation. The same is true on the extremely-low data 1\% data percentage mark, where ELECTRA has degraded by 15.04\% on average, which is 6.2\% higher than BERT's average degradation of 8.84\%.

\begin{figure}[htp]
    \centering
    \includegraphics[width=0.48\textwidth]{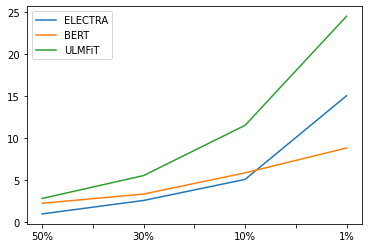}
    \caption{\label{fig:degradation-curve-models} 
    Per-technique degradation curves, averaging the performance of all models belonging to one technique. ULMFiT still remains the easiest model to degrade as the number of training examples reduce. ELECTRA starts to degrade strongly after the 10\% mark, while BERT remains to degrade slowly.}
\end{figure}

In extremely low-data domains, we see that BERT is more resilient to performance degradation than ELECTRA is. ELECTRA is shown to degrade exponentially as the number of training examples is reduced. As shown in Figure \ref{fig:degradation-curve-models}, while BERT's degradation on average remains relatively linear, ELECTRA starts degrading faster and faster as we approach the 1\% (4,200 examples) data percentage mark. When looking at degradation speeds, it is also evident that ELECTRA degrades more (average degradation speed of 5.46) while BERT degrades less (average degradation speed of 2.53).


We hypothesize that this is a direct effect of their pretraining schemes. ELECTRA is trained without a specific downstream task in mind, while BERT is trained considering sentence-pair classification tasks, leading to its use of next sentence prediction as a secondary pretraining task. Since BERT's biases are more adjusted to sentence-pair classification, we can hypothesize that it should perform reliably well even when finetuned with little data, as it already has an ``idea'' of how to perform the task.

\begin{figure}[htp]
    \centering
    \includegraphics[width=0.48\textwidth]{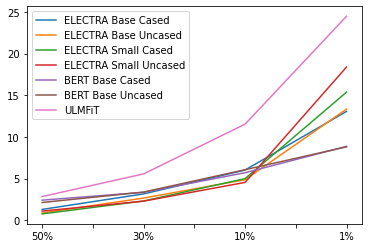}
    \caption{\label{fig:degradation-curve} 
    Per-model degradation curves. ULMFiT degrades the fastest out of the three transfer learning techniques. ELECTRA begins to degrade at the 10\% data percentage mark. Both ELECTRA and ULMFiT degrade exponentially, while BERT degrades nearly linearly, which is likely due to it's pretraining scheme designed for sentence-pair classification tasks.}
\end{figure}

In terms of per-model degradation, while the small ELECTRA models outperformed their larger base counterparts in the full dataset, we show that the base models are more resilient to degradation. As shown in Figure \ref{fig:degradation-curve}, this is more evident as we approach the 1\% data percentage mark, with the small uncased model degrading 5.04\% more than the base uncased model, and the small cased model degrading 2.31\% more than its base cased counterpart. In terms of speed, we also see that the small models degrade more (average degradation speed of 6.31) than the base models (average degradation speed of 4.62).


\subsection{Heuristics for Choosing Techniques for Low-Data Domains}
Overall, the finetuning results and the degradation tests give us good heuristics for choosing which techniques are appropriate for different tasks when dealing with low-data domains.

ELECTRA is most effective in the general use-case. When there is a lot of task-specific data to finetune a model, we see that ELECTRA is more effective than the baselines BERT and ULMFiT. ELECTRA is also best to use when there is little pretraining data available, as is with the case of the low-resource language Filipino. Since there is less pretraining data, ELECTRA's more data-efficient pretraining scheme will impart more learned priors than BERT's masked language modeling objective will. From our results, we hypothesize that the same will be true when compared with other Transformer-based pretraining techniques that use the masked language modeling objective, such as RoBERTa \cite{liu2019roberta}.

However, while ELECTRA is effective in the general case, this does not mean that BERT will be deprecated anytime soon. BERT is very effective in the low-data case, especially in tasks that deal with sentence-pair classification such as natural language inference and sentence entailment. Since BERT's pretraining scheme is designed with sentence-pair classification tasks in mind, it will perform well for such tasks even with little finetuning data as it already has an idea how to perform these tasks due to its pretraining. As we show with empirical results, BERT also degrades slower than ELECTRA, and should be more robust for various tasks in low-data domains and low-resource languages.

While both Transformer-based finetuning techniques outperform ULMFiT in the degradation tests, this does not mean that RNN-based methods do not have a use in current research dealing with language inference tasks. On the full dataset, we see that ULMFiT performed with accuracy comparable to BERT, albeit degrading the fastest on average on the degradation tests. ULMFiT's fast degradation is likely due to it being RNN-based, which has significantly less representational capacity than the larger Transformers that leverage attention used in the study. While this is the case, in settings where there is enough data to finetune, ULMFiT (and other RNN-based transfer learning techniques) will perform comparably to the Transformer-based techniques when tuned properly. ULMFiT's AWD-LSTM \cite{merity2017regularizing} backbone also enjoys the benefit of being cheaper and faster to train. In cases where there is a lack of resources to use Transformers effectively, RNN-based models will still suffice, assuming there is an abundance of data.

\section{Conclusion}
In this paper, we proposed an automatic method for creating sentence entailment benchmark datasets using news articles. Through our method, datasets can be generated quickly and cost-efficiently, while ensuring that they are challenging enough to accurately benchmark performance of high capacity models. In addition, our method leverages the abundance of news articles online, which allows datasets even in low-resource languages to be created.

Using our method, we produce the first sentence entailment benchmark dataset in Filipino which we call NewsPH-NLI. We also produce pretrained Transformers based on the ELECTRA pretraining scheme, which we benchmark on our dataset against two widely-used techniques, BERT and ULMFiT.

We shed light on the true performance of transfer learning techniques when operating in low-data domains to solve a hard task. We show the importance of the choice of pretraining task to the effectiveness of a Transformer when finetuned with little data. We also show that while newer techniques outperform older established ones, they may still perform worse when dealing with low-resource languages.

For future work, we recommend further studies on automatic corpus generation be done; particularly on correctness checking. The biggest disadvantage that our method has is that to fully ensure correctness, humans will still have to evaluate the resulting dataset. Should an automatic technique to verify correctness be made, our dataset generation method will be more robust, and can then be adapted to generate other tasks that require more human supervision in creating, such as summarization and translation.

%
%
%
\bibliographystyle{splncs04}
\bibliography{bibliography.bib}

\begin{thebibliography}{10}
\providecommand{\url}[1]{\texttt{#1}}
\providecommand{\urlprefix}{URL }
\providecommand{\doi}[1]{https://doi.org/#1}

\bibitem{bowman2015large}
Bowman, S.R., Angeli, G., Potts, C., Manning, C.D.: A large annotated corpus
  for learning natural language inference. arXiv preprint arXiv:1508.05326
  (2015)

\bibitem{canavilhas2007web}
Canavilhas, J.: Web journalism: from the inverted pyramid to the tumbled
  pyramid. Biblioteca on-line de ci{\^e}ncias da comunica{\c{c}}{\~a}o  (2007)

\bibitem{clark2020electra}
Clark, K., Luong, M.T., Le, Q.V., Manning, C.D.: Electra: Pre-training text
  encoders as discriminators rather than generators. arXiv preprint
  arXiv:2003.10555  (2020)

\bibitem{conneau2018xnli}
Conneau, A., Rinott, R., Lample, G., Williams, A., Bowman, S.R., Schwenk, H.,
  Stoyanov, V.: Xnli: Evaluating cross-lingual sentence representations. In:
  Proceedings of the 2018 Conference on Empirical Methods in Natural Language
  Processing. Association for Computational Linguistics (2018)

\bibitem{cruz2019evaluating}
Cruz, J.C.B., Cheng, C.: Evaluating language model finetuning techniques for
  low-resource languages. arXiv preprint arXiv:1907.00409  (2019)

\bibitem{cruz2020establishing}
Cruz, J.C.B., Cheng, C.: Establishing baselines for text classification in
  low-resource languages. arXiv preprint arXiv:2005.02068  (2020)

\bibitem{cruz2020localization}
Cruz, J.C.B., Tan, J.A., Cheng, C.: Localization of fake news detection via
  multitask transfer learning. In: Proceedings of The 12th Language Resources
  and Evaluation Conference. pp. 2596--2604 (2020)

\bibitem{currey2019incorporating}
Currey, A., Heafield, K.: Incorporating source syntax into transformer-based
  neural machine translation. In: Proceedings of the Fourth Conference on
  Machine Translation (Volume 1: Research Papers). pp. 24--33 (2019)

\bibitem{devlin2018bert}
Devlin, J., Chang, M.W., Lee, K., Toutanova, K.: Bert: Pre-training of deep
  bidirectional transformers for language understanding. arXiv preprint
  arXiv:1810.04805  (2018)

\bibitem{hinds1977paragraph}
Hinds, J.: Paragraph structure and pronominalization. Paper in Linguistics
  \textbf{10}(1-2),  77--99 (1977)

\bibitem{hoey2008beginning}
Hoey, M., O'Donnell, M.B.: The beginning of something important? corpus
  evidence on the text beginnings of hard news stories. Corpus Linguistics,
  Computer Tools and Applications: State of the Art. PALC  \textbf{2007},
  189--212 (2008)

\bibitem{howard2020fastai}
Howard, J., Gugger, S.: Fastai: A layered api for deep learning. Information
  \textbf{11}(2), ~108 (2020)

\bibitem{howard2018universal}
Howard, J., Ruder, S.: Universal language model fine-tuning for text
  classification. arXiv preprint arXiv:1801.06146  (2018)

\bibitem{khandelwal2019sample}
Khandelwal, U., Clark, K., Jurafsky, D., Kaiser, L.: Sample efficient text
  summarization using a single pre-trained transformer. arXiv preprint
  arXiv:1905.08836  (2019)

\bibitem{kingma2014adam}
Kingma, D.P., Ba, J.: Adam: A method for stochastic optimization. arXiv
  preprint arXiv:1412.6980  (2014)

\bibitem{lamble2013news}
Lamble, S.: News as it happens: An introduction to journalism. Oxford
  University Press (2013)

\bibitem{le2014distributed}
Le, Q., Mikolov, T.: Distributed representations of sentences and documents.
  In: International conference on machine learning. pp. 1188--1196 (2014)

\bibitem{liu2019roberta}
Liu, Y., Ott, M., Goyal, N., Du, J., Joshi, M., Chen, D., Levy, O., Lewis, M.,
  Zettlemoyer, L., Stoyanov, V.: Roberta: A robustly optimized bert pretraining
  approach. arXiv preprint arXiv:1907.11692  (2019)

\bibitem{merity2017regularizing}
Merity, S., Keskar, N.S., Socher, R.: Regularizing and optimizing lstm language
  models. arXiv preprint arXiv:1708.02182  (2017)

\bibitem{murray2019auto}
Murray, K., Kinnison, J., Nguyen, T.Q., Scheirer, W., Chiang, D.: Auto-sizing
  the transformer network: Improving speed, efficiency, and performance for
  low-resource machine translation. arXiv preprint arXiv:1910.06717  (2019)

\bibitem{myagmar2019cross}
Myagmar, B., Li, J., Kimura, S.: Cross-domain sentiment classification with
  bidirectional contextualized transformer language models. IEEE Access
  \textbf{7},  163219--163230 (2019)

\bibitem{norambuena2020evaluating}
Norambuena, B.K., Horning, M., Mitra, T.: Evaluating the inverted pyramid
  structure through automatic 5w1h extraction and summarization. In: Proc. of
  the 2020 Computation+ Journalism Symposium. Computation+ Journalism. pp.~1--7
  (2020)

\bibitem{po2003news}
Po{\"{}}~ttker, H.: News and its communicative quality: The inverted
  pyramid—when and why did it appear? Journalism Studies  \textbf{4}(4),
  501--511 (2003)

\bibitem{smith2017cyclical}
Smith, L.N.: Cyclical learning rates for training neural networks. In: 2017
  IEEE Winter Conference on Applications of Computer Vision (WACV). pp.
  464--472. IEEE (2017)

\bibitem{vaswani2017attention}
Vaswani, A., Shazeer, N., Parmar, N., Uszkoreit, J., Jones, L., Gomez, A.N.,
  Kaiser, {\L}., Polosukhin, I.: Attention is all you need. In: Advances in
  neural information processing systems. pp. 5998--6008 (2017)

\bibitem{velasco2020pagsusuri}
Velasco, D.J.: Pagsusuri ng rnn-based transfer learning technique sa
  low-resource language. arXiv preprint arXiv:2010.06447  (2020)

\bibitem{williams2017broad}
Williams, A., Nangia, N., Bowman, S.R.: A broad-coverage challenge corpus for
  sentence understanding through inference. arXiv preprint arXiv:1704.05426
  (2017)

\end{thebibliography}

\end{document}